\documentclass[11pt,a4paper]{article}
\usepackage[hyperref]{acl2017}
\usepackage{times}
\usepackage{latexsym}
\usepackage{graphicx}

\usepackage{url}

\aclfinalcopy 

\title{Word Embedding for Response-To-Text Assessment of Evidence}

\author{Haoran Zhang \\
  Department of Computer Science\\
  University of Pittsburgh \\
  Pittsburgh, PA 15260 \\
  {\tt colinzhang@cs.pitt.edu} \\\And
  Diane Litman \\
  Department of Computer Science \& LRDC\\
 University of Pittsburgh \\
  Pittsburgh, PA 15260 \\
  {\tt litman@cs.pitt.edu} \\}

\date{}

\begin{document}
\maketitle
\begin{abstract}
Manually grading the Response to Text Assessment (RTA) is labor intensive. Therefore, an automatic method is being developed for scoring analytical writing when the RTA is administered in large numbers of classrooms. Our long-term goal is to also use this scoring method to provide formative feedback to students and teachers about students’ writing quality. As a first step towards this goal, interpretable features for automatically scoring the evidence rubric of the RTA have been developed. In this paper, we present a simple but promising method for improving evidence scoring by employing the word embedding model. We evaluate our method on corpora of responses written by upper elementary students.
\end{abstract}

\section{Introduction}
In \citet{correnti2013assessing}, it was noted that the 2010 Common Core State Standards  emphasize the ability of young students from grades 4-8 to interpret and evaluate texts, construct logical arguments based on substantive claims, and marshal relevant evidence in support of these claims.   \citet{correnti2013assessing} relatedly developed the Response to Text Assessment (RTA) for assessing students' analytic response-to-text writing skills. The RTA was designed to evaluate writing skills in Analysis, Evidence, Organization, Style, and MUGS (Mechanics, Usage, Grammar, and Spelling) dimensions. 
To both score the RTA and provide formative feedback to students and teachers at scale, an automated RTA scoring tool is now being developed \citep{rahimi2017assessing}. 

This paper focuses on the Evidence dimension of the RTA, which evaluates students' ability to find and use evidence from an article to support their position. \citet{rahimi2014automatic} previously developed a set of interpretable features for scoring the Evidence rubric of RTA. Although these features significantly improve over competitive baselines, the feature extraction approach is largely based on lexical matching and can be enhanced.

The contributions of this paper are as follows. First, we employ a new way of using the word embedding model to enhance the system of \citet{rahimi2014automatic}. Second, 
we use word embeddings to deal with  
noisy data given the disparate writing skills of students at the upper elementary level.

In the following sections, we first present research on related topics, describe our corpora, and review  the interpretable features developed by \citet{rahimi2014automatic}. Next, we explain how we use the word embedding model for feature extraction to improve performance by addressing the limitations of prior work. Finally, we discuss the results of our experiments and present  future plans.

\section{Related Work}
Most research studies in automated essay scoring have focused on holistic rubrics \citep{shermis2003automated,attali2006automated}. In contrast, our work focuses on evaluating a single dimension to obtain a rubric score for students' use of evidence from a source text to support their stated position. To evaluate the content of students' essays, ~\citet{louis2010off} presented a method to detect if an essay is off-topic. \citet{xie2012exploring} presented a method to evaluate content features by measuring the similarity between essays. \citet{burstein2001enriching} and \citet{ong2014ontology} both presented methods to use argumentation mining techniques to evaluate the students' use of evidence to support claims in persuasive essays. However, those studies are different from this work in that they did not measure how the essay uses material from the source article. Furthermore, young students find it difficult to use sophisticated argumentation structure in their essays. 

\citet{rahimi2014automatic} presented a set of interpretable rubric features that measure the relatedness between students' essays and a source article by extracting evidence from the students' essays. However, evidence from students' essays could not always be extracted by their word matching method. There are some potential solutions using the word embedding model. \citet{rei2016sentence} presented a method to evaluate topical relevance by estimating sentence similarity using weighted-embedding. \citet{kenter2015short} evaluated short text similarity with word embedding. \citet{kiela2015specializing} developed specialized word embedding by employing external resources. However, none of these methods address highly noisy essays written by young students. 

\section{Data}
Our response-to-text essay corpora were all collected from classrooms using the following procedure. The teacher first read aloud a text while students followed along with their copy. After the teacher explained some predefined vocabulary and discussed standardized questions at designated points, there is a prompt at the end of the text which asks students to write an essay in response to the prompt. Figure~\ref{fig:exampleessay} shows the prompt of $RTA_{MVP}$

Two forms of the RTA have been developed, based on different articles that students read before writing essays in response to a prompt. The first form is $RTA_{MVP}$ and is based on an article from \emph{Time for Kids} about the Millennium Villages Project, an effort by the United Nations to end 
poverty in a rural village in Sauri, Kenya. The other form is $RTA_{Space}$, based on a developed article about the importance of space exploration. Below is a small excerpt from the $RTA_{MVP}$ article. Evidence from the text that expert human graders want to see in students' essays are in bold.

\begin{quote}
``Today, Yala Sub-District {\bf Hospital has medicine}, {\bf free of charge}, {\bf for all of the most common diseases}. {\bf Water is connected to the hospital}, which also has a {\bf generator for electricity}. {\bf Bed nets are used} in every sleeping site in Sauri.''
\end{quote}

\begin{table}[h]
\begin{center}
\begin{tabular}{|r|ccc|}
\hline  & \bf $Space$ & \bf $MVP_L$ & \bf $MVP_H$  \\ \hline
\bf  Score 1 & 538 & 535 & 317 \\
& (26\%) & (30\%) & (27\%) \\
\bf Score 2 & 789 & 709 & 488 \\
& (38\%) & (39\%) & (42\%) \\
\bf Score 3 & 512 & 374 & 242 \\
& (25\%) & (21\%) & (21\%) \\
\bf Score 4 & 237 & 186 & 119 \\
& (11\%) & (10\%) & (10\%) \\ \hline
\bf Total & 2076 & 1804 & 1166 \\ 
\bf Double-Rated & 2076 & 847 & 1156 \\ \hline
\bf Kappa & 0.338 & 0.490 & 0.479 \\
\bf QWKappa & 0.651 & 0.775 & 0.734 \\
\hline
\end{tabular}
\end{center}
\caption{\label{distribution} The  distribution of Evidence scores, and grading agreement of two raters.}
\end{table}

Two corpora of $RTA_{MVP}$ from lower and higher age groups were introduced in \citet{correnti2013assessing}. One group included grades 4-6 (denoted by $MVP_L$), and the other group included grades 6-8 (denoted by $MVP_H$). The students in each age group represent different levels of writing proficiency. We also combined these two corpora to form a larger corpus, denoted by $MVP_{ALL}$. The corpus of the $RTA_{Space}$ is collected only from students of grades 6-8 (denoted by $Space$). 

Based on the rubric criterion shown in Table~\ref{rubricTable}, the essays in each corpus were annotated by two raters on a scale of 1 to 4,  from low to high. Raters are experts and trained undergraduates. Table~\ref{distribution} shows the distribution of Evidence scores from the first rater and the agreement (Kappa, and Quadratic Weighted Kappa) between two raters of the double-rated portion. All experiment performances will be measured by Quadratic Weighted Kappa between the score from prediction and the first rater. The reason to only use the score of the first rater is that the first rater graded more essays. Figure~\ref{fig:exampleessay} shows an essay with a score of 3.

\begin{table*}
\centering
\scalebox{0.6}{
\begin{tabular}{|p{3.75cm}p{5cm}p{5cm}p{5cm}p{5cm}|}
\hline & \bf 1 & \bf 2 & \bf 3 & \bf 4 \\
\hline Number of Pieces of evidence & Features one or no pieces of evidence (NPE) & Features at least 2 pieces of evidence (NPE) & Features at least 3 pieces of evidence (NPE) & Features at least 3 pieces of evidence (NPE) \\
Relevance of evidence & Selects inappropriate or irrelevant details from the text to support key idea (SPC); references to text feature serious factual errors or omissions & Selects some appropriate and relevant evidence to support key idea, or evidence is provided for some ideas, but not actually the key idea (SPC); evidence may contain a factual error or omission & Selects pieces of evidence from the text that are appropriate and relevant to key idea (SPC) & Selects evidence from the text that clearly and effectively supports key idea \\
Specificity of evidence & Provides general or cursory evidence from the text (SPC) & Provides general or cursory evidence from the text (SPC) & Provides specific evidence from the text (SPC) & Provides pieces of evidence that are detailed and specific (SPC) \\
Elaboration of Evidence & Evidence may be listed in a sentence (CON) & Evidence provided may be listed in a sentence, not expanded upon (CON) & Attempts to elaborate upon evidence (CON) & Evidence must be used to support key idea / inference(s) \\
Plagiarism & Summarize entire text or copies heavily from text (in these cases, the response automatically receives a 1) & & & \\

\hline
\end{tabular}}
\caption{\label{rubricTable} Rubric for the Evidence dimension of RTA. The abbreviations in the parentheses identify the corresponding feature group discussed in the Rubric Features section of this paper that is aligned with that specific criteria \citep{rahimi2017assessing}.} 
\end{table*}

\begin{figure}[h]
\begin{quote}
{\bf Prompt: } The author provided one specific example of how the quality of life can be improved by the Millennium Villages Project in Sauri, Kenya. Based on the article, did the author provide a convincing argument that winning the fight against poverty is achievable in our lifetime? Explain why or why not with 3-4 examples from the text to support your answer.

{\bf Essay: }In my opinion I think that they will achieve it in lifetime. During the years threw {\bf 2004 and 2008 they made progress}. People didn’t have the money to buy the stuff in 2004. {\em The hospital was packed with patients\/} and they didn’t have alot of treatment in 2004. In 2008 it changed the {\em hospital had medicine\/}, {\em free of charge\/}, and {\bf for all the common dieases}. {\em Water was connected to the hospital\/} and has a {\em generator for electricity\/}. {\bf Everybody has net} in their site. {\em The hunger crisis has been addressed\/} with {\bf fertilizer and seeds}, as well as the {\em tools needed to maintain the food\/}. {\em The school has no fees\/} and {\em they serve lunch\/}. To me that’s sounds like it is going achieve it in the lifetime.
\end{quote}
\caption{The prompt of $RTA_{MVP}$ and an example essay with score of 3.}
\label{fig:exampleessay}
\end{figure}


\section{Rubric Features}
Based on the rubric criterion for the evidence dimension, \citet{rahimi2014automatic} developed a set of interpretable features. By using this set of features, a predicting model can be trained for automated essay scoring in the evidence dimension.

{\bf Number of Pieces of Evidence (NPE):} A good essay should mention evidence from the article as much as possible. To extract the NPE feature, they manually craft a topic word list based on the article. Then, they use a simple window-based algorithm with a fixed size window to extract this feature. If a window contains at least two words from the topic list, they consider this window to contain evidence related to a topic. To avoid redundancy, each topic is only counted once. Words from the window and crafted list will only be considered a match if they are exactly the same. This feature is an integer to represent the number of topics that are mentioned by the essay.

{\bf Concentration (CON):} Rather than list all the topics in the essay, a good essay should explain each topic with details. The same topic word list and simple window-based algorithm are used for extracting the CON feature. An essay is concentrated if the essay has fewer than 3 sentences that mention at least one of the topic words. Therefore, this feature is a binary feature. The value is 1 if the essay is concentrated, otherwise it is 0.

{\bf Specificity (SPC):} A good essay should use relevant examples as much as possible. For matching SPC feature, experts manually craft an example list based on the article. Each example belongs to one topic, and is an aspect of a specific detail about the topic. For each example, the same window-based algorithm is used for matching. If the window contains at least two words from an example, they consider the window to mention this example. 
Therefore, the SPC feature is an integer vector. 
Each value in the vector represents how many examples in this topic were mentioned by the essay. To avoid redundancy, each example is only to be counted at most one time. 
The length of the vector is the same as the number of categories of examples in the crafted list. 

{\bf Word Count (WOC):} The SPC feature can capture how many evidences were mentioned in the essay, but it cannot represent if these pieces of evidence support key ideas effectively. From previous work, we know longer essays tend to have higher scores. Thus, they use word count as a potentially helpful fallback feature. This feature is an integer.

\section{Word Embedding Feature Extraction} 
Based on the results of \citet{rahimi2014automatic}, the interpretable rubric-based features outperform competitive baselines. However, 
there are limitations in 
their feature extraction method. It cannot extract all examples mentioned by the essay due to the use of simple exact matching.

First, students use their own vocabularies other than words in the crafted list. For instance, some students use the word ``power'' instead of ``electricity'' from the crafted list.

Second, according to our corpora, students at the upper elementary level make spelling mistakes, and sometimes they make mistakes in the same way. For example, around 1 out of 10 students misspell ``poverty'' as ``proverty'' instead. Therefore, evidence with student spelling mistakes cannot be extracted. However, the evidence dimension of RTA does not penalize students for misspelling words. \citet{rahimi2014automatic} showed that manual spelling corrections indeed improves performance, but not significantly.

Finally, tenses used by students can sometimes be different from that of the article. Although a stemming algorithm can solve this problem, sometimes there are words that slip through the process. For example, ``went'' is the past tense of ``go'', but stemming would miss this conjugation. Therefore, ``go'' and ``went'' would not be considered a match.

To address the limitations above, we introduced the Word2vec (the skip-gram (SG) and the continuous bag-of-words (CBOW)) word embedding model presented by \citet{mikolov2013efficient} into the feature extraction process. By mapping words from the vocabulary to vectors of real numbers, the similarity between two words can be calculated. Words with high similarity can be considered a match. Because words in the same context tend to have similar meaning, they would therefore have higher similarity.


We use the word embedding model as a supplement to the original feature extraction process, and use the same searching window algorithm presented by \citet{rahimi2014automatic}. If a word in a student's essay is not exactly the same as the word in the crafted list, the cosine similarity between these two words is calculated by the word embedding model. We consider them matching, if the similarity is higher than a threshold. 

In Figure~\ref{fig:exampleessay}, the phrases in italics are examples extracted by the existing feature extraction method. For instance, ``water was connected to the hospital'' can be found because ``water'' and ``hospital'' are exactly the same as words in the crafted list. However, ``for all the common dieases'' cannot be found due to misspelling of ``disease''. Additional examples that can be extracted by the word embedding model are in bold.

\section{Experimental Setup}
We configure experiments to test several hypotheses: H1) the model with the word embedding trained on our own corpus will outperform or at least perform equally well as the baseline (denoted by $Rubric$) presented by \citet{rahimi2014automatic}. H2) the model with the word embedding trained on our corpus will outperform or at least perform equally well as the model with off-the-shelf word embedding models. H3) the model with word embedding trained on our own corpus will generalize better across students of different ages. Note that while all models with word embeddings use the same features as the $Rubric$ baseline, the feature extraction process was changed to allow non-exact matching via the word embeddings.

We stratify each corpus into 3 parts: 40\% of the data are used for training the word embedding models; 20\% of the data are used to select the best word embedding model and best threshold (this is the development set of our model); and another 40\% of data are used for final testing.

For word embedding model training, we also add essays not graded by the first rater ($Space$ has 229, $MVP_L$ has 222, $MVP_H$ has 296, and $MVP_{ALL}$ has 518) to 40\% of the data from the corpus in order to enlarge the training corpus to get better word embedding models. We train multiple word embedding models with different parameters, and select the best word embedding model by using the development set.

Two off-the-shelf word embeddings are used for comparison. \citet{mikolov2013distributed} presented vectors that have 300 dimensions and were trained on a newspaper corpus of about 100 billion words. The other is presented by \citet{baroni2014don} and includes 400 dimensions, with the context window size of 5, 10 negative samples and subsampling. 

We use 10 runs of 10-fold cross validation in the final testing, with Random Forest (max-depth = 5) implemented in Weka \citep{witten2016data} as the classifier. This is the setting used by \citet{rahimi2014automatic}. Since our corpora are imbalanced with respect to the four evidence scores being predicted (Table~\ref{distribution}), we use SMOTE oversampling method \citep{chawla2002smote}. This involves creating ``synthetic'' examples for minority classes. We only oversample the training data. All experiment performances are measured by Quadratic Weighted Kappa (QWKappa).

\section{Results and Discussion}
We first examine H1. The results shown in Table~\ref{cvperfomance} partially support this hypothesis. The skip-gram embedding yields a higher performance or performs equally well as the rubric baseline on most corpora, except for $MVP_H$. The skip-gram embedding significantly improves performance for the lower grade corpus. Meanwhile, the skip-gram embedding is always significantly better than the continuous bag-of-words embedding. 

Second, we examine H2. Again, the results shown in Table~\ref{cvperfomance} partially support this hypothesis. The skip-gram embedding trained on our corpus outperform Baroni's embedding on $Space$ and $MVP_L$. While Baroni's embedding is significantly better than the skip-gram embedding on $MVP_H$ and $MVP_{ALL}$.

\begin{table*}
\centering
\begin{tabular}{|l|c|cc|cc|}
\hline & & \multicolumn{2}{c|}{Off-the-Shelf} & \multicolumn{2}{c|}{On Our Corpus} \\
\bf Corpus & \bf Rubric(1) & \bf Baroni(2) &\bf Mikolov(3) & \bf SG(4) & \bf CBOW(5) \\ \hline
$Space$ & 0.606(2) & 0.594 & 0.606(2) & \bf 0.611(2,5) & 0.600(2)\\
$MVP_L$ & 0.628 & 0.666(1,3,5) & 0.623 & \bf 0.682(1,2,3,5) & 0.641(1,3)\\
$MVP_H$ & \bf 0.599(3,4,5) & 0.593(3,4,5) & 0.582(5) & 0.583(5) & 0.556\\
$MVP_{ALL}$ & 0.624(5) & \bf 0.645(1,3,4,5) & 0.634(1,5) & 0.634(1,5) &0.614\\
\hline
\end{tabular}
\caption{\label{cvperfomance} The performance (QWKappa) of the off-the-shelf embeddings and embeddings trained on our corpus compared to the rubric baseline on all corpora. The numbers in parenthesis show the model numbers over which the current model performs significantly better. The best results in each row are in bold.} 
\end{table*}

Third, we examine H3, by training models from one corpus and testing it on 10 disjointed sets of the other test corpus. We do it 10 times and average the results in order to perform significance testing. The results shown in Table~\ref{ccperfomance} support this hypothesis. The skip-gram word embedding model outperform all other models. 

\begin{table*}
\centering
\begin{tabular}{|ll|c|cc|cc|}
\hline & & & \multicolumn{2}{c|}{Off-the-Shelf} & \multicolumn{2}{c|}{On Our Corpus} \\
\bf Train & \bf Test & \bf Rubric(1) & \bf Baroni(2) &\bf Mikolov(3) & \bf SG(4) & \bf CBOW(5) \\ \hline
$MVP_L$ & $MVP_H$ & 0.582(3) & 0.609 (1,3,5) & 0.555 & \bf 0.615(1,2,3,5) & 0.596(1,3)\\
$MVP_H$ & $MVP_L$ & 0.604 & 0.629(1,3,5) & 0.620(1,5) & \bf 0.644(1,2,3,5) & 0.605\\
\hline
\end{tabular}
\caption{\label{ccperfomance} The performance (QWKappa) of the off-the-shelf embeddings and embeddings trained on our corpus compared to the rubric baseline. The numbers in parenthesis show the model numbers over which the current model performs significantly better. The best results in each row are in bold.} 
\end{table*}

As we can see, the skip-gram embedding outperforms the continuous bag-of-words embedding in all experiments. One possible reason for this is that the skip-gram is better than the continuous bag-of-words for infrequent words \citep{mikolov2013distributed}. In the continuous bag-of-words, vectors from the context will be averaged before predicting the current word, while the skip-gram does not. Therefore, it remains a better representation for rare words. Most students tend to use words that appear directly from the article, and only a small portion of students introduce their own vocabularies into their essays. Therefore, the word embedding is good with infrequent words and tends to work well for our purposes. 

In examining the performances of the two off-the-shelf word embeddings, Mikolov's embedding cannot help with our task, because it has less preprocessing of its training corpus. Therefore, the embedding is case sensitive and contains symbols and numbers. For example, it matches ``2015'' with ``000''. Furthermore, its training corpus comes from newspapers, which may contain more high-level English that students may not use, and professional writing has few to no spelling mistakes. Although Baroni's embedding also has no spelling mistakes, it was trained on a corpus containing more genres of writing and has more preprocessing. Thus, it is a better fit to our work compared to Mikolov's embedding. 

In comparing the performance of the skip-gram embedding and Baroni's embedding, there are many differences. First, even though the skip-gram embedding partially solves the tense problem, Baroni's embedding solves it better because it has a larger training corpus. Second, the larger training corpus contains no or significantly fewer spelling mistakes, and therefore it cannot solve the spelling problem at all. On the other hand, the skip-gram embedding solves the spelling problem better, because it was trained on our own corpus. For instance, it can match ``proverty'' with ``poverty'', while Baroni's embedding cannot. Third, the skip-gram embedding cannot address a vocabulary problem as well as the Baroni's embedding because of the small training corpus. Baroni's embedding matches ``power'' with ``electricity'', while the skip-gram embedding does not. Nevertheless, the skip-gram embedding still partially addresses this problem, for example, it matches ``mosquitoes'' with ``malaria'' due to relatedness. Last, Baroni's embedding was trained on a corpus that is thousands of times larger than our corpus. However, it does not address our problems significantly better than the skip-gram embedding due to generalization. In contrast, our task-dependent word embedding is only trained on a small corpus while outperforming or at least performing equally well as Baroni's embedding.

Overall, the skip-gram embedding tends to find examples by implicit relations. For instance, ``winning against poverty possible achievable lifetime'' is an example from the article and in the meantime the prompt asks students ``Did the author provide a convincing argument that winning the fight against poverty is achievable in our lifetime?''. Consequently, students may mention this example by only answering ``Yes, the author convinced me.''. However, the skip-gram embedding can extract this implicit example.

\section{Conclusion and Future Work}
We have presented several simple but promising uses of the word embedding method that improve evidence scoring in corpora of responses to texts  written by upper elementary students. In our results, a task-dependent word embedding model trained on our small corpus was the most helpful in improving the baseline model. However, the word embedding model still measures additional information that is not necessary in our work. Improving the word embedding model or the feature extraction process is thus our most likely future endeavor.

One potential improvement is re-defining the loss function of the word embedding model, since the word embedding measures not only the similarity between two words, but also the relatedness between them. However, our work is not helped by matching related words too much. For example, we want to match ``poverty'' with ``proverty'', while we do not want to match ``water'' with ``electricity'', even though students mention them together frequently. Therefore, we could limit this measurement by modifying the loss function of the word embedding. \citet{kiela2015specializing} presented a specialized word embedding by employing an external thesaurus list. However, it does not fit to our task, because the list contains high-level English words that will not be used by young students.

Another area for future investigation is improving the word embedding models trained on our corpus. Although they improved performance, they were trained on a corpus from one form of the RTA and tested on the same RTA. Thus, another possible improvement is generalizing the modelfrom one RTA to another RTA.

\section*{Acknowledgments}
We would like to show our appreciation to every member of the RTA group for sharing their pearls of wisdom with us. We are also immensely grateful to Dr. Richard Correnti, Deanna Prine, and Zahra Rahimi for their comments on an earlier version of the paper.

The research reported here was supported, in whole or in part, by the Institute of Education Sciences, U.S. Department of Education, through Grant R305A160245 to the University of Pittsburgh. The opinions expressed are those of the authors and do not represent the views of the Institute or the U.S. Department of Education.

\bibliography{acl2017}

\begin{thebibliography}{}
\expandafter\ifx\csname natexlab\endcsname\relax\def\natexlab#1{#1}\fi

\bibitem[{Attali and Burstein(2006)}]{attali2006automated}
Yigal Attali and Jill Burstein. 2006.
\newblock Automated essay scoring with e-rater{\textregistered} v. 2.
\newblock {\em The Journal of Technology, Learning and Assessment\/} 4(3).

\bibitem[{Baroni et~al.(2014)Baroni, Dinu, and Kruszewski}]{baroni2014don}
Marco Baroni, Georgiana Dinu, and Germ{\'a}n Kruszewski. 2014.
\newblock Don't count, predict! a systematic comparison of context-counting vs.
  context-predicting semantic vectors.
\newblock In {\em ACL (1)\/}. pages 238--247.

\bibitem[{Burstein et~al.(2001)Burstein, Kukich, Wolff, Lu, and
  Chodorow}]{burstein2001enriching}
Jill Burstein, Karen Kukich, Susanne Wolff, Chi Lu, and Martin Chodorow. 2001.
\newblock Enriching automated essay scoring using discourse marking. .

\bibitem[{Chawla et~al.(2002)Chawla, Bowyer, Hall, and
  Kegelmeyer}]{chawla2002smote}
Nitesh~V Chawla, Kevin~W Bowyer, Lawrence~O Hall, and W~Philip Kegelmeyer.
  2002.
\newblock Smote: synthetic minority over-sampling technique.
\newblock {\em Journal of artificial intelligence research\/} 16:321--357.

\bibitem[{Correnti et~al.(2013)Correnti, Matsumura, Hamilton, and
  Wang}]{correnti2013assessing}
Richard Correnti, Lindsay~Clare Matsumura, Laura Hamilton, and Elaine Wang.
  2013.
\newblock Assessing students' skills at writing analytically in response to
  texts.
\newblock {\em The Elementary School Journal\/} 114(2):142--177.

\bibitem[{Kenter and de~Rijke(2015)}]{kenter2015short}
Tom Kenter and Maarten de~Rijke. 2015.
\newblock Short text similarity with word embeddings.
\newblock In {\em Proceedings of the 24th ACM International on Conference on
  Information and Knowledge Management\/}. ACM, pages 1411--1420.

\bibitem[{Kiela et~al.(2015)Kiela, Hill, and Clark}]{kiela2015specializing}
Douwe Kiela, Felix Hill, and Stephen Clark. 2015.
\newblock Specializing word embeddings for similarity or relatedness.
\newblock In {\em EMNLP\/}. pages 2044--2048.

\bibitem[{Louis and Higgins(2010)}]{louis2010off}
Annie Louis and Derrick Higgins. 2010.
\newblock Off-topic essay detection using short prompt texts.
\newblock In {\em Proceedings of the NAACL HLT 2010 Fifth Workshop on
  Innovative Use of NLP for Building Educational Applications\/}. Association
  for Computational Linguistics, pages 92--95.

\bibitem[{Mikolov et~al.(2013{\natexlab{a}})Mikolov, Chen, Corrado, and
  Dean}]{mikolov2013efficient}
Tomas Mikolov, Kai Chen, Greg Corrado, and Jeffrey Dean. 2013{\natexlab{a}}.
\newblock Efficient estimation of word representations in vector space.
\newblock {\em arXiv preprint arXiv:1301.3781\/} .

\bibitem[{Mikolov et~al.(2013{\natexlab{b}})Mikolov, Sutskever, Chen, Corrado,
  and Dean}]{mikolov2013distributed}
Tomas Mikolov, Ilya Sutskever, Kai Chen, Greg~S Corrado, and Jeff Dean.
  2013{\natexlab{b}}.
\newblock Distributed representations of words and phrases and their
  compositionality.
\newblock In {\em Advances in neural information processing systems\/}. pages
  3111--3119.

\bibitem[{Ong et~al.(2014)Ong, Litman, and Brusilovsky}]{ong2014ontology}
Nathan Ong, Diane Litman, and Alexandra Brusilovsky. 2014.
\newblock Ontology-based argument mining and automatic essay scoring.
\newblock In {\em Proceedings of the First Workshop on Argumentation Mining\/}.
  pages 24--28.

\bibitem[{Rahimi et~al.(2017)Rahimi, Litman, Correnti, Wang, and
  Matsumura}]{rahimi2017assessing}
Zahra Rahimi, Diane Litman, Richard Correnti, Elaine Wang, and Lindsay~Clare
  Matsumura. 2017.
\newblock Assessing students’ use of evidence and organization in
  response-to-text writing: Using natural language processing for rubric-based
  automated scoring.
\newblock {\em International Journal of Artificial Intelligence in Education\/}
  pages 1--35.

\bibitem[{Rahimi et~al.(2014)Rahimi, Litman, Correnti, Matsumura, Wang, and
  Kisa}]{rahimi2014automatic}
Zahra Rahimi, Diane~J Litman, Richard Correnti, Lindsay~Clare Matsumura, Elaine
  Wang, and Zahid Kisa. 2014.
\newblock Automatic scoring of an analytical response-to-text assessment.
\newblock In {\em International Conference on Intelligent Tutoring Systems\/}.
  Springer, pages 601--610.

\bibitem[{Rei and Cummins(2016)}]{rei2016sentence}
Marek Rei and Ronan Cummins. 2016.
\newblock Sentence similarity measures for fine-grained estimation of topical
  relevance in learner essays.
\newblock {\em arXiv preprint arXiv:1606.03144\/} .

\bibitem[{Shermis and Burstein(2003)}]{shermis2003automated}
Mark~D Shermis and Jill~C Burstein. 2003.
\newblock {\em Automated essay scoring: A cross-disciplinary perspective\/}.
\newblock Routledge.

\bibitem[{Witten et~al.(2016)Witten, Frank, Hall, and Pal}]{witten2016data}
Ian~H Witten, Eibe Frank, Mark~A Hall, and Christopher~J Pal. 2016.
\newblock {\em Data Mining: Practical machine learning tools and techniques\/}.
\newblock Morgan Kaufmann.

\bibitem[{Xie et~al.(2012)Xie, Evanini, and Zechner}]{xie2012exploring}
Shasha Xie, Keelan Evanini, and Klaus Zechner. 2012.
\newblock Exploring content features for automated speech scoring.
\newblock In {\em Proceedings of the 2012 Conference of the North American
  Chapter of the Association for Computational Linguistics: Human Language
  Technologies\/}. Association for Computational Linguistics, pages 103--111.

\end{thebibliography}
\bibliographystyle{acl_natbib}

\end{document}